%% file: 1st_sub_main_Arxiv_all.tex
\documentclass[10pt,twocolumn,letterpaper]{article}

\usepackage[pagenumbers]{cvpr} 

\usepackage{booktabs,multirow}
\input{preamble}
\definecolor{cvprblue}{rgb}{0.21,0.49,0.74}
\usepackage[pagebackref,breaklinks,colorlinks,allcolors=cvprblue]{hyperref}

\def\confName{CVPR}
\def\confYear{2027}

\title{JEPA-Tuner: From Predictive Fine-Tuning to Asynchronous Execution of Vision-Language-Action Models}

\author{
Jingxiang Qu$^{1,*}$ \quad
Tian Xie$^{2,*}$ \quad
Youngjin Hong$^{2}$ \quad
Changhyun Choi$^{2}$ \\
Yang Katie Zhao$^{2,\dagger}$ \quad
Yi Liu$^{1,\dagger}$ \\[0.5em]
$^{1}$Stony Brook University $^{2}$University of Minnesota -- Twin Cities  \\
{\small $^{*}$Equal contribution.
\quad $^{\dagger}$Corresponding authors.}
}
\begin{document}
\maketitle
\begingroup
\renewcommand{\thefootnote}{}
\footnotetext{Contacts:
\textit{jiqqu@cs.stonybrook.edu}
and
\textit{yi.liu.4@stonybrook.edu.}}
\endgroup
\input{sec/0_abstract}
\vspace{-6pt}
\input{sec/1_intro}
\input{sec/2_Related_Works}
\input{sec/3_Preliminaries}
\input{sec/4_Method}

\input{sec/5_Experiments_Meta-World-short}
\input{sec/6_Conclusions}
\input{sec/7_LLM_Usage_Statement}
{
    \small
    \bibliographystyle{ieeenat_fullname}
    \bibliography{main}
}

\clearpage
\appendix  

\renewcommand{\theHsection}{supp.\Alph{section}}

\twocolumn[{%
  \centering
  {\LARGE\bfseries Supplementary Materials\par}
  \vspace{1.5em}
}]

\input{sec/JEPA_Tuner_Supplement}

\end{document}

%% file: sec/0_abstract.tex
\begin{abstract}
Vision-Language-Action (VLA) models generate sequences of actions from visual observations, proprioceptive states, and language instructions. In embodied interaction, executing these actions influences subsequent observations and robot states, providing a natural predictive signal for policy learning. However, standard action-supervised fine-tuning does not explicitly capture these action-conditioned temporal dependencies in the guidance consumed by the action predictor. We introduce \textbf{JEPA-Tuner}, a predictive fine-tuning framework that regularizes the VLA's native policy guidance through action-conditioned \textbf{future} prediction. During fine-tuning, a predictor in JEPA-Tuner learns to match guidance encoded from a future observation given the current guidance and committed actions. This objective encourages a cross-modal guidance adapter to capture how the scene and robot state evolve under those actions, while jointly training the action expert on both observed and predicted guidance keeps the representation informative for control. In addition, when deploying the VLA model with asynchronous action-chunk inference, the pre-trained predictor will anticipate the guidance needed at the next chunk boundary. The policy can thus generate the next chunk in advance and execute it from its first action without overlap correction or additional predictor training, enabling efficient and high-quality asynchronous execution. Experiments on LIBERO and Meta-World show that JEPA-Tuner improves task success under synchronous inference with better multi-modal guidance embeddings. For asynchronous control, JEPA-Tuner improves performance and smoothness across different latency settings on both LIBERO simulator and multi-size real robot demonstrations. The project website with video demos is at \url{https://qujx.github.io/JEPA-Tuner-Website/}.
\end{abstract}

%% file: sec/1_intro.tex
\vspace{-10pt}
\section{Introduction}
\label{sec:introduction}
\vspace{-4pt}

\leavevmode
{\clubpenalty=0\relax
Vision-Language-Action (VLA) models integrate visual observations, language instructions, and proprioceptive states to generate robot action chunks~\citep{openvla,pi0,pi05,smolvla}. Many recent VLAs couple pretrained vision-language models (VLMs)~\citep{qwen,paligemma} with action experts through multimodal policy guidance (vision, language, robotic states, etc.). Structuring this multimodal guidance is important for both policy adaptation and asynchronous execution. When a robot executes an action on an object, it changes both the arm configuration and the object's position. Prior works used this relationship to anticipate input to a policy. For instance, VLASH~\citep{vlash} estimates future proprioceptive states using current states and committed actions, while FutureRTC~\citep{futurertc} predicts visual representations and proprioceptive states to correct a frozen VLA policy. These examples suggest that multimodal control signals are coupled through learnable predictive dependencies. However, standard action-supervised fine-tuning does not enforce these dependencies in the guidance consumed by the action expert. We therefore use the observations of executed actions as an additional training signal for the guidance that conditions the action expert. Specifically, we encourage \emph{action-conditioned predictive consistency} by jointly training a predictor to predict future guidance and proprioception from the current guidance, and further align the policy on it. \par}

Action-conditioned prediction also offers a natural way to address inference latency in asynchronous control. As Fig.~\ref{fig:rtc_challenges} illustrates, the robot continues to move while the next chunk is computed, shifting execution toward later predictions conditioned on an earlier observation. Real-Time Chunking (RTC)~\citep{rtc} corrects the overlap with committed actions through inference-time inpainting, while training based method (e.g., T-RTC~\citep{training_time_rtc}, Legato~\citep{legato}) learn smoothness during training to continue the committed trajectory. Although conditioning on the committed action prefix can smooth transitions between chunks, a strong bias toward continuing the previous trajectory can limit the action expert's ability to respond to state changes, reducing closed-loop task success under inference delay~\citep{legato,soft_rtc}.


To capture action-conditioned dependencies in multimodal guidance and better balance trajectory smoothness with action accuracy during asynchronous inference, we introduce \textbf{JEPA-Tuner}, a fine-tuning framework inspired by joint-embedding predictive architectures for robot control~\citep{jepa_vla,vla_jepa,slim}. Specifically, a cross-modal adapter mixes visual, language, and state information at the VLA's native input prefix. Given the adapted guidance at step $t$, the robot state, and the next $d$ demonstrated actions, a predictor estimates the future guidance and proprioceptive state at step $t+d$. In VLA model finetuning, we further train the action expert to generate action chunks from both observed and predicted guidance, while jointly training the predictor to match the guidance encoded from the observation at $t+d$. This joint objective encourages the predicted guidance to capture the future scene and state while retaining the information needed for action generation. During asynchronous inference, we reuse the same predictor with the actions committed before the handoff, allowing the expert to generate the next chunk in advance and execute it from its first action without overlap correction.

\noindent Our main contributions are as follows:
\begin{itemize}
    \item We identify action-conditioned temporal dependencies in multimodal guidance that standard action-supervised VLA fine-tuning ignored. Motivated by this observation, we propose \textbf{JEPA-Tuner}, which improves VLA fine-tuning by regularizing native multimodal guidance through future prediction while jointly training the action expert on both observed and predicted guidance.
    
    \item To address latency-induced conditioning mismatch during asynchronous control, we develop a predictive execution scheme that reuses the predictor learned by JEPA-Tuner during fine-tuning to anticipate guidance at the next chunk handoff. This enables each new chunk to directly execute from its first action with anticipated guidance without additional training or overlap correction.
    
    \item Experiments on LIBERO and Meta-World demonstrate improved VLA fine-tuning performance. Asynchronous evaluations further demonstrate a better trade-off between task success and action smoothness, both in simulation and on robot platforms of different scales, SO-101 and ALOHA, under standard and severe inference latency.
\end{itemize}

%% file: sec/2_Related_Works.tex
\vspace{-4pt}
\section{Related Work}
\label{sec:related_work}
\vspace{-4pt}
\subsection{Vision-Language-Action Models}
\vspace{-4pt}
Vision-Language-Action (VLA) models and generalist robot policies integrate visual observations and language instructions for robot control. RT-2~\citep{rt2} represents actions as tokens, while OpenVLA~\citep{openvla} and Octo~\citep{octo} advance generalist policy pretraining and adaptation. OpenVLA-OFT~\citep{openvla_oft} studies efficient action decoding and fine-tuning. Recently, diffusion~\citep{diffusion} and flow matching~\citep{flow_matching} play an important role in generative modeling across multiple data modalities~\citep{sdxl, gaga, p2flow}. These frameworks also support continuous action generation: $\pi_0$~\citep{pi0} introduces a flow-matching action expert, $\pi_{0.5}$~\citep{pi05} extends this paradigm through heterogeneous co-training, and SmolVLA~\citep{smolvla} emphasizes compact, affordable action-chunk policies. Unless otherwise specified, we consider VLA policies that generate continuous action chunks using diffusion or flow matching.

\vspace{-4pt}
\subsection{Predictive Learning for VLA Policies}
\vspace{-4pt}
VLA-JEPA~\citep{vla_jepa} and JEPA-WAM~\citep{JEPA-WAM} impose temporal supervision through visual targets defined by pretrained V-JEPA encoders, while JEPA-VLA~\citep{jepa_vla} augments existing VLAs with pretrained video-predictive embeddings. WoG~\citep{WoG} learns compact future-observation conditions for action generation, and SLIM~\citep{slim} models action-conditioned transitions within its own learned observation-latent space. Their future-representation targets or transferred features are primarily visual, even when actions and proprioception provide conditioning. These formulations do not address the joint problem of shaping an existing VLA's native multimodal guidance and forecasting that guidance at execution boundaries under inference latency. 

JEPA-Tuner addresses this joint problem by learning action-conditioned predictive consistency within an existing VLA's native policy guidance space. As a plug-in fine-tuning framework, it couples future visual-guidance and proprioceptive-state prediction with action supervision, directly regularizing the temporal evolution of the conditioning consumed by the action expert. Crucially, the same learned predictor is reused to anticipate guidance at the next execution boundary, unifying downstream representation adaptation with latency-aware asynchronous control.

\vspace{-4pt}
\subsection{Asynchronous Inference of VLA Models}
\vspace{-4pt}
Asynchronous VLA methods seek continuous, temporally aligned control under inference latency. RTC~\citep{rtc} uses inference-time inpainting conditioned on committed actions, while Training-Time RTC~\citep{training_time_rtc} learns committed-prefix conditioning during training. Legato~\citep{legato} similarly learns continuation for flow-based policies. These methods promote temporal compatibility in action space.

Another complementary strategy anticipates execution-time conditioning. VLASH~\citep{vlash} rolls proprioception forward while retaining earlier visual observations; EagleVLA~\citep{eaglevla} predicts future compressed VLM features; and FutureRTC~\citep{futurertc} forecasts visual latents and proprioceptive states for a frozen policy. Latent Bridge~\citep{latent_bridge} predicts conditioning-feature changes to reduce backbone invocations. However, these methods primarily exploit prediction for deployment, leaving underexplored whether action-conditioned predictive regularization of native multimodal guidance can improve VLA fine-tuning. JEPA-Tuner obtains two benefits from a single fine-tuning procedure: improved VLA finetuning performance through predictive guidance regularization and a predictor directly reusable for asynchronous control. The jointly learned predictor anticipates execution-boundary guidance, supporting prefix execution without a separate predictor-training stage or inference-time overlap correction.

%% file: sec/3_Preliminaries.tex
\vspace{-2pt}
\section{Preliminaries}
\label{sec:Preliminaries}
\vspace{-2pt}
\begin{figure*}[t]
    \centering
    \vspace{-14pt}\includegraphics[width=\textwidth]{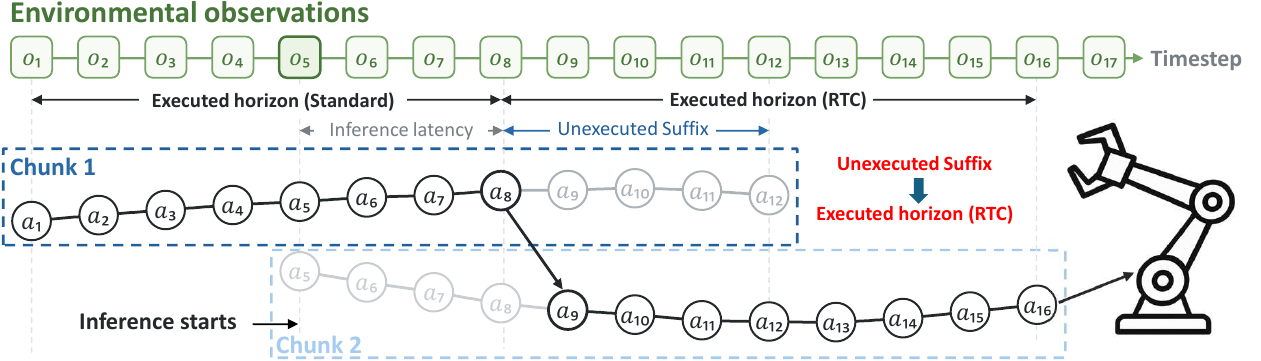}
    \vspace{-20pt}
    \caption{\textbf{Latency-induced discrepancy in asynchronous execution.}
    With chunk size $H=12$, execution horizon $E=8$, and inference latency $\Delta=4$, inference for Chunk~2 starts from $o_5$ while Chunk~1 executes $a_5,\ldots,a_8$. Chunk~2's elapsed prefix is skipped, and its predictions $a_9,\ldots,a_{16}$ are executed. Relative to executing its first $E$ predictions, RTC shifts the execution window toward later predictions conditioned on $o_5$, using part of the normally unexecuted suffix.}
    \label{fig:rtc_challenges}
    \vspace{-14pt}
\end{figure*}

\subsection{Vision-Language-Action Models}

Continuous-action VLAs, including $\pi_0$~\citep{pi0}, $\pi_{0.5}$~\citep{pi05}, and SmolVLA~\citep{smolvla}, condition a generative action expert on visual observations, proprioceptive states, and language instructions. Given visual observation $o_t$, robot state $s_t$, and instruction $\ell$, we write the policy guidance as
\begin{equation}
g_t = E_{\theta}(o_t,s_t,\ell),
\label{eq:vla_guidance}
\end{equation}
where $E_{\theta}$ maps the inputs to a native conditioning interface and $g_t$ denotes the resulting \emph{policy guidance}, serving as the conditions in action experts. The interface depends on the backbone and consists of multimodal prefix embeddings. The downstream mapping $D_{\theta}$ includes subsequent multimodal processing and the action expert, generating a chunk of $H$ actions:
\begin{equation}
\mathbf{A}_{t:t+H-1}
= (a_t,\ldots,a_{t+H-1})
\sim D_{\theta}(\cdot\mid g_t).
\label{eq:vla_action_chunk}
\end{equation}
Diffusion and Flow matching-based policies iteratively transform noise into actions while keeping $g_t$ fixed. Standard receding-horizon execution uses the first $E\leq H$ actions and discards the remaining $H-E$ actions, termed the \emph{unexecuted suffix}, before replanning from new observations. 

\subsection{Asynchronous VLA Inference}

Asynchronous inference predicts next action chunk during robot execution~\citep{rtc,smolvla}. Suppose inference starts at time $t$ using guidance $g_t$, with end-to-end latency $\Delta$ measured in control steps. During this interval, the robot executes committed actions $\mathbf{A}^{\mathrm{com}}_{t:t+\Delta-1}$. When the new chunk becomes available at $t+\Delta$, its observation-derived guidance remains $g_t$, although the robot and environment have evolved. This temporal mismatch constitutes \emph{prediction--execution misalignment}~\citep{vlash,eaglevla,futurertc}, as shown in Fig.~\ref{fig:rtc_challenges}.

For a chunk indexed relative to observation time $t$, the first $\Delta$ action positions correspond to timesteps that have elapsed by inference completion. Temporal alignment therefore skips this prefix and executes positions $\Delta,\ldots,\Delta+E-1$, provided that $\Delta+E\leq H$. The resulting window contains later predictions while retaining the observation-derived conditioning from time $t$.

RTC~\citep{rtc} handles asynchronous continuation through inference-time inpainting conditioned on committed actions, promoting continuity across chunks. Figure~\ref{fig:rtc_challenges} illustrates $H=12$, $E=8$, and $\Delta=4$: inference from $o_5$ runs while Chunk~1 executes $a_5,\ldots,a_8$, after which Chunk~2 supplies $a_9,\ldots,a_{16}$. Its executed window includes $a_{13},\ldots,a_{16}$ from the suffix omitted under first-$E$ execution. Although inpainting reconciles overlapping actions, these predictions retain visual conditioning from $o_5$. Consequently, latency shifts execution toward later predictions under earlier observations, while iterative overlap correction introduces additional computation.

%% file: sec/4_Method.tex
\vspace{-4pt}
\section{JEPA-Tuner Methodology}
\label{sec:method}
\begin{figure*}[t]
    \centering
    \vspace{-20pt}
    \includegraphics[width=\textwidth]{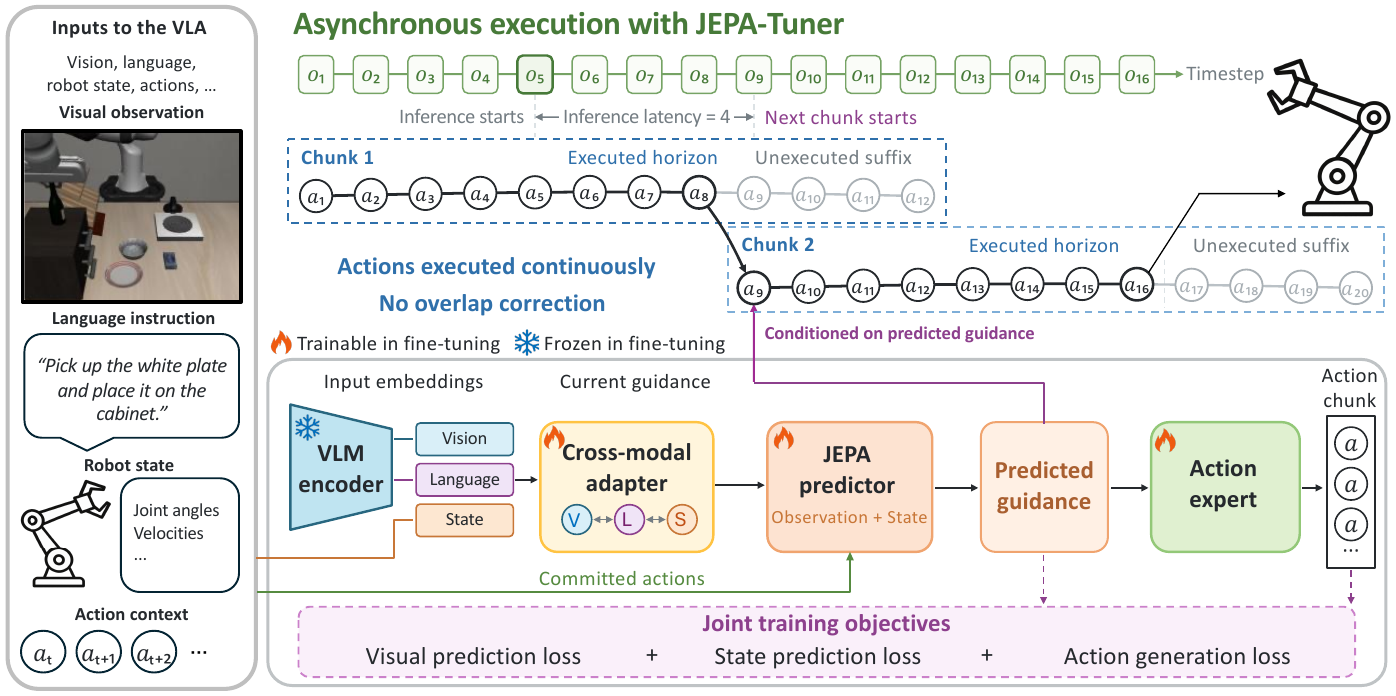}
    \vspace{-22pt}
    \caption{\textbf{Predictive fine-tuning and asynchronous execution with JEPA-Tuner. During fine-tuning (bottom)}, the cross-modal adapter, JEPA predictor, and action expert are trained jointly while the VLM encoder stays frozen. The adapter fuses visual, language, and state embeddings into policy guidance. Given this guidance, the current robot state, and $d$ intervening demonstrated actions, the predictor estimates visual guidance and robot state at $t+d$. The encoded future observation and measured future state supervise these predictions, while demonstrated action chunks supervise the expert under both observed and predicted guidance. \textbf{During asynchronous execution (top)}, the same predictor anticipates guidance at the next chunk boundary. For instance, with chunk size $H=12$, execution horizon $E=8$, and latency $\Delta=4$, inference starts from $o_5$ and uses the committed actions $a_5,\ldots,a_8$ to predict guidance for step~$9$. The expert decodes Chunk~2 from this guidance. Its first eight actions, $a_9,\ldots,a_{16}$, follow the Chunk~1 immediately without overlap correction.
    }
    \label{fig:jepa_tuner}
    \vspace{-16pt}
\end{figure*}

\vspace{-2pt}
In this section, we first identify two challenges: underexploited action-conditioned multimodal dependencies in VLA fine-tuning and stale guidance during asynchronous execution. We then introduce \textbf{JEPA-Tuner}, which addresses both through predictive modeling in the VLA's native guidance space, as introduced in Fig.~\ref{fig:jepa_tuner}. 
\vspace{-2pt}
\subsection{Challenges}
\label{sec:challenges}
\vspace{-2pt}
\paragraph{Underexploited intrinsic multimodal dependencies in VLA fine-tuning.}
Visual observations, language instructions, proprioceptive states, and actions describe interconnected aspects of robotic control. Vision and proprioception describe the environment and robot configuration, while language specifies task intent. Together, they condition action generation. Executed actions subsequently change the robot state and observed environment. These complementary relationships suggest that policy guidance should capture both how multimodal context informs actions and how that context evolves under their execution. However, standard action-supervised fine-tuning primarily optimizes observation-to-action prediction, without explicitly modeling the latter dependency. Guidance sufficient for action generation is therefore not necessarily structured to predict its future counterparts. This motivates \emph{action-conditioned predictive consistency} in the VLA's native guidance space, encouraging representations that preserve intrinsic temporal dependencies while remaining informative for control.

\vspace{-14pt}
\paragraph{High-latency execution under stale conditioning.}
Chunked policies typically execute only the first $E<H$ actions before replanning from updated observations, leaving an unexecuted suffix of length $H-E$~\citep{pi0,diffusionpolicy}. Under asynchronous inference, latency separates the observation used for prediction from the context in which the resulting actions take effect. As discussed in Sec.~\ref{sec:Preliminaries}, skipping the elapsed prefix shifts the executed positions from $0,\ldots,E-1$ to $\Delta,\ldots,\Delta+E-1$, provided that $\Delta+E\leq H$. Increasing latency consequently brings more of the normally unexecuted suffix into the execution window while retaining earlier observation-derived guidance. For example, with $H=12$, $E=8$, and $\Delta=4$ in Fig.~\ref{fig:rtc_challenges}, four of the eight executed predictions come from this suffix. This misalignment brings severe discontinuity in asynchronous execution.

To overcome such discontinuity, RTC~\citep{rtc} requires iterative inference-time correction to reconcile overlapping chunks, while training-based methods such as T-RTC~\citep{training_time_rtc} and Legato~\citep{legato} learn committed-prefix continuation, which may bias action generation toward the preceding trajectory. These methods nevertheless exhibit substantial performance degradation under high latency, as shown by our LIBERO results in Sec.~\ref{sec:experiments}. Motivated by these limitations, we seek a new asynchronous execution strategy that achieves a better trade-off between trajectory smoothness and task success.

\vspace{-4pt}
\subsection{Fine-Tuning VLAs via JEPA Modeling}
\label{sec:jepa_finetuning}
\paragraph{Adapting native multimodal guidance.}
Following the decomposition in Sec.~\ref{sec:Preliminaries}, we instantiate policy guidance at the VLA's native prefix interface, before its multimodal transformer layers. Let $R_{\theta}$ denote the policy's input embedding map and $T_{\theta}$ a trainable cross-modal adapter:
\begin{equation}
r_t=R_{\theta}(o_t,s_t,\ell),
\qquad
g_t=T_{\theta}(r_t).
\label{eq:jepa_adapted_guidance}
\end{equation}
The adapter uses residual attention across visual, language, and state representations, allowing each component to incorporate information from the others while preserving the native conditioning structure. The downstream mapping $D_{\theta}$ comprises the remaining multimodal processing and action expert. Predictive regularization thus acts on the policy's own guidance, and action supervision requires that guidance to remain useful for control.
\vspace{-12pt}
\paragraph{Action-conditioned guidance prediction.}
A Joint-Embedding Predictive Architecture (JEPA) learns representations through prediction in embedding space. We apply this principle to paired policy contexts separated by $d$ control steps. Future observations and robot states define target guidance in the same native conditioning space:
\begin{equation}
\bar{g}_{t+d}
=
E_{\bar{\theta}}(o_{t+d},s_{t+d},\ell),
\label{eq:jepa_target_guidance}
\end{equation}
where $E_{\bar{\theta}}$ is the target counterpart of $E_{\theta}=T_{\theta}\circ R_{\theta}$, sharing its conditioning interface and frozen pretrained components. Let $\bar{V}_{t+d}$ denote its visual component. Given current guidance, robot state, and demonstrated intervening actions, the predictor estimates future visual guidance and proprioception:
\begin{equation}
\begin{aligned}
(\widehat{V}_{t+d},\widehat{s}_{t+d})
=
P_{\phi}\!\left(
g_t,s_t,\mathbf{A}^{\mathrm{demo}}_{t:t+d-1},d
\right).
\end{aligned}
\label{eq:jepa_predictor}
\end{equation}
Here, visual guidance refers to the visual positions within the adapted prefix, which already incorporate cross-modal context. The instruction remains task conditioning rather than a prediction target.

The predicted components are expressed through the backbone's native interface to construct future guidance:
\begin{equation}
\widehat{g}_{t+d}
=
C_{\theta}\!\left(
\widehat{V}_{t+d},\widehat{s}_{t+d};
r_t,\ell
\right).
\label{eq:jepa_guidance_composition}
\end{equation}
The composition map $C_{\theta}$ retains the predicted visual guidance, supplies the predicted state through the policy's state-conditioning map, and preserves the instruction context. The current prefix $r_t$ provides context for this construction. Consequently, observed and anticipated guidance share the conditioning structure expected by the same downstream policy.
\vspace{-12pt}
\paragraph{Predictive consistency.}
We supervise visual evolution in guidance space and robot dynamics in state space:
\begin{equation}
\mathcal{L}_{\mathrm{JEPA}}
=
\mathbb{E}_{t,d}\!\left[
\ell_g(\widehat{V}_{t+d},\bar{V}_{t+d})
+
\lambda_s\ell_s(\widehat{s}_{t+d},s_{t+d})
\right],
\label{eq:jepa_component_losses}
\end{equation}
where $\ell_g$ combines normalized squared error and cosine distance, $\ell_s$ is normalized state-prediction error, and $\lambda_s$ balances the two components. Their separate supervision preserves the distinct scales and semantics of visual and proprioceptive information while coupling their prediction through shared guidance and intervening actions.
\vspace{-12pt}
\paragraph{Joint predictive and action supervision.}
Embedding agreement alone does not ensure that predicted guidance supports accurate actions. We therefore supervise action generation under both observed and anticipated guidance:
\begin{equation}
\begin{aligned}
\mathcal{L}_{\mathrm{act}}
&=
\mathbb{E}_{t}\!\left[
\ell_{\mathrm{act}}\!\left(
g_t,\mathbf{A}^{\mathrm{demo}}_{t:t+H-1}
\right)\right]
\\
&\quad+
\lambda_{\mathrm{dec}}\mathbb{E}_{t,d}\!\left[
\ell_{\mathrm{act}}\!\left(
\widehat{g}_{t+d},
\mathbf{A}^{\mathrm{demo}}_{t+d:t+d+H-1}
\right)\right],
\end{aligned}
\label{eq:jepa_action_objective}
\end{equation}
where $\ell_{\mathrm{act}}$ is the backbone's action-learning objective and $\lambda_{\mathrm{dec}}$ weights future-action supervision. For flow-based policies, both terms use the standard flow-matching loss against their corresponding demonstrated chunks. The second term makes future prediction relevant to control by requiring the anticipated guidance to condition actions appropriate to the future context.

The resulting objective is
\begin{equation}
\min_{\theta,\phi}\;
\mathcal{L}_{\mathrm{act}}
+
\lambda_{\mathrm{JEPA}}\mathcal{L}_{\mathrm{JEPA}}.
\label{eq:jepa_joint_objective}
\end{equation}
We jointly optimize the guidance adapter, predictor, action expert, and applicable native state-conditioning parameters while keeping the pretrained visual and language backbone frozen. This couples temporal predictability with action generation in a single fine-tuning procedure. It also enables the policy to use the anticipated conditioning supplied during asynchronous deployment.
\vspace{-4pt}
\subsection{From Predictive Learning to Asynchronous Execution}
\label{sec:jepa_async}
\vspace{-2pt}
\paragraph{Anticipating execution-time guidance.}
The predictive relationship learned during fine-tuning directly matches the temporal structure of asynchronous execution. During training, demonstrated actions connect the current context to its future target. At deployment, committed policy actions define the interval until the next execution boundary. JEPA-Tuner therefore reuses the jointly learned predictor to anticipate that boundary's guidance, without a separate predictor-training stage.

Let inference begin at time $t$ with latency $\Delta$, and set the prediction horizon to $d=\Delta$. While the robot executes committed actions, the next chunk is generated as
\begin{equation}
\begin{aligned}
(\widehat{V}_{t+d},\widehat{s}_{t+d})
&=
P_{\phi}\!\left(
g_t,s_t,\mathbf{A}^{\mathrm{com}}_{t:t+d-1},d
\right),
\\
\widehat{g}_{t+d}
&=
C_{\theta}\!\left(
\widehat{V}_{t+d},\widehat{s}_{t+d};r_t,\ell
\right),
\\
\widehat{\mathbf{A}}_{t+d:t+d+H-1}
&\sim
D_{\theta}(\cdot\mid\widehat{g}_{t+d}).
\end{aligned}
\label{eq:jepa_async_prediction}
\end{equation}
The chunk becomes available at $t+d$ and is indexed from that execution boundary. For a sequential inference schedule whose requests lie within the preceding execution window, continuous execution requires
\begin{equation}
\Delta=d\leq E\leq H,
\label{eq:jepa_feasibility}
\end{equation}
with committed actions covering the inference interval.
\vspace{-12pt}
\paragraph{Execution alignment and computational benefit.}
Each new chunk executes its first $E$ actions, leaving the remaining $H-E$ actions as an unexecuted suffix. Anticipated guidance therefore preserves the standard prefix-execution pattern without skipping an elapsed prefix or correcting overlapping predictions. A forward pass through the jointly learned predictor replaces inference-time overlap correction; the policy still generates a full chunk once per execution window. JEPA-Tuner consequently provides both a regularized policy and an anticipatory control mechanism, connecting fine-tuning with asynchronous deployment.

%% file: sec/5_Experiments_Meta-World-short.tex
\vspace{-6pt}
\section{Experiments}
\label{sec:experiments}
\vspace{-2pt}
We first use synchronous inference to test whether action-conditioned predictive supervision improves VLA fine-tuning. We follow the standard protocols on LIBERO~\citep{libero} and Meta-World MT50~\citep{metaworld}, counting an episode as successful when the policy achieves the task goal within the episode limit. We then reuse the predictor during asynchronous execution and evaluate both task success and the smoothness of action trajectories, first on LIBERO and then on real robots SO-101 and ALOHA. The supplementary material gives the backbone-specific implementations, baseline configurations, deployment settings, latency measurements, and ablation studies on the multimodal adapter and predictor in JEPA-Tuner. Specifically, the ablations show that our design enables accurate prediction in the guidance embedding space without compromising action learning, while multimodal interaction itself can also improves policy learning. Therefore, JEPA-Tuner combines guidance prediction with multimodal interaction to improve VLA fine-tuning and enable asynchronous execution by anticipating the guidance required at the next chunk handoff.

\vspace{-4pt}
\subsection{Benefits for VLA Fine-Tuning}
\label{sec:exp_finetuning}
\vspace{-2pt}
We compare JEPA-Tuner with standard fine-tuning (Standard FT) on SmolVLA~\citep{smolvla}, $\pi_{0.5}$~\citep{pi05}, and GR00T~\citep{gr00t}. During synchronous evaluation, both policies decode actions from the current observation, and JEPA-Tuner's predictor is inactive. On LIBERO, we use the published Standard FT results~\citep{vla_jepa} and fine-tune the unavailable weights by ourselves. On Meta-World MT50, we match demonstrations and training budgets within each backbone and evaluate 50 episodes per task.

\begin{table}[t]
    \centering
    \caption{\textbf{Synchronous success on LIBERO~\citep{libero}} (\%). Avg gives equal weight to each suite.}
    \vspace{-10pt}\label{tab:libero_finetuning}
    \footnotesize
    \setlength{\tabcolsep}{2pt}
    \renewcommand{\arraystretch}{1.05}
    \begin{tabular*}{\columnwidth}{@{\extracolsep{\fill}}lccccc@{}}
        \toprule
        Method & Spatial & Object & Goal & LIBERO-10 & Avg $\uparrow$ \\
        \midrule
        UniVLA~\citep{univla}
        & 96.5 & 96.8 & 95.6 & 92.0 & 95.2 \\
        OpenVLA-OFT~\citep{openvla_oft}
        & 97.6 & 98.4 & 97.9 & 94.5 & 97.1 \\
        $\pi_0$~\citep{pi0}
        & 96.8 & 98.8 & 95.8 & 85.2 & 94.2 \\
        $\pi_0$-FAST~\citep{pi0fast}
        & 96.4 & 96.8 & 88.6 & 60.2 & 85.5 \\
        WorldVLA~\citep{worldvla}
        & 87.6 & 96.2 & 83.4 & 60.0 & 81.8 \\
        villa-X~\citep{villa_x}
        & 97.5 & 97.0 & 91.5 & 74.5 & 90.1 \\
        VLA-JEPA~\citep{vla_jepa}
        & 96.2 & 99.6 & 97.2 & 95.8 & 97.2 \\
        \midrule
        Standard FT (SmolVLA)~\citep{smolvla}
        & 78.6 & 88.6 & 87.8 & 56.6 & 77.9 \\
        Standard FT ($\pi_{0.5}$)~\citep{pi05}
        & 98.8 & 98.2 & 98.0 & 92.4 & 96.9 \\
        Standard FT (GR00T)~\citep{gr00t}
        & 94.4 & 97.6 & 93.0 & 90.6 & 93.9 \\
        \midrule
        JEPA-Tuner (SmolVLA)
        & 77.2 & 93.0 & 88.6 & 54.8 & 78.4 \\
        JEPA-Tuner ($\pi_{0.5}$)
        & 98.4 & 98.6 & 97.8 & 95.6 & 97.6 \\
        JEPA-Tuner (GR00T)
        & 96.6 & 99.8 & 97.8 & 93.0 & 96.8 \\
        \bottomrule
    \end{tabular*}
    \vspace{-12pt}
\end{table}

\begin{table}[t]
    \centering
    \caption{\textbf{Synchronous success on Meta-World MT50~\citep{metaworld}} (\%). Avg weights each difficulty group by its number of tasks.}
    \vspace{-10pt}
    \label{tab:metaworld_finetuning}
    \footnotesize
    \setlength{\tabcolsep}{1pt}
    \renewcommand{\arraystretch}{1.05}
    \begin{tabular*}{\columnwidth}{@{\extracolsep{\fill}}lccccc@{}}
        \toprule
        Method & Easy & Medium & Hard & Very Hard & Avg $\uparrow$ \\
        \midrule
        Diffusion Policy~\citep{diffusionpolicy}
        & 23.1 & 10.7 & 1.9 & 6.1 & 16.1 \\
        TinyVLA-H~\citep{tinyvla}
        & 77.6 & 21.5 & 11.4 & 15.8 & 51.1 \\
        $\pi_0$~\citep{pi0}
        & 71.8 & 48.2 & 41.7 & 30.0 & 58.8 \\
        \midrule
        Standard FT (SmolVLA)
        & 81.0 & 46.7 & 37.7 & 42.8 & 64.4 \\
        Standard FT ($\pi_{0.5}$)
        & 54.0 & 18.2 & 8.7 & 19.2 & 37.2 \\
        Standard FT (GR00T)
        & 76.3 & 30.4 & 34.0 & 42.0 & 57.7 \\
        \midrule
        JEPA-Tuner (SmolVLA)
        & 80.9 & 50.2 & 38.7 & 50.0 & 66.0 \\
        JEPA-Tuner ($\pi_{0.5}$)
        & 64.0 & 19.3 & 12.0 & 24.8 & 44.0 \\
        JEPA-Tuner (GR00T)
        & 80.3 & 44.2 & 47.0 & 52.0 & 65.5 \\
        \bottomrule
    \end{tabular*}
    \vspace{-16pt}
\end{table}

JEPA-Tuner improves average task success over Standard FT for all three VLA backbones on both LIBERO and Meta-World MT50 (Tables~\ref{tab:libero_finetuning} and~\ref{tab:metaworld_finetuning}). On LIBERO, for example, average success rises from $77.9\%$ to $78.4\%$ for SmolVLA and from $93.9\%$ to $96.8\%$ for GR00T. On Meta-World MT50, $\pi_{0.5}$ improves from $37.2\%$ to $44.0\%$, and GR00T improves from $57.7\%$ to $65.5\%$. In all cases, the action expert decodes directly from observed guidance with the predictor disabled at inference. These results suggest that incorporating action-conditioned prediction of future guidance into VLA fine-tuning helps the policy generate more effective action chunks from current observations, improving manipulation success even when latency compensation is unnecessary.

\begin{figure*}[t]
    \centering
    \vspace{-18pt}\includegraphics[width=\textwidth]{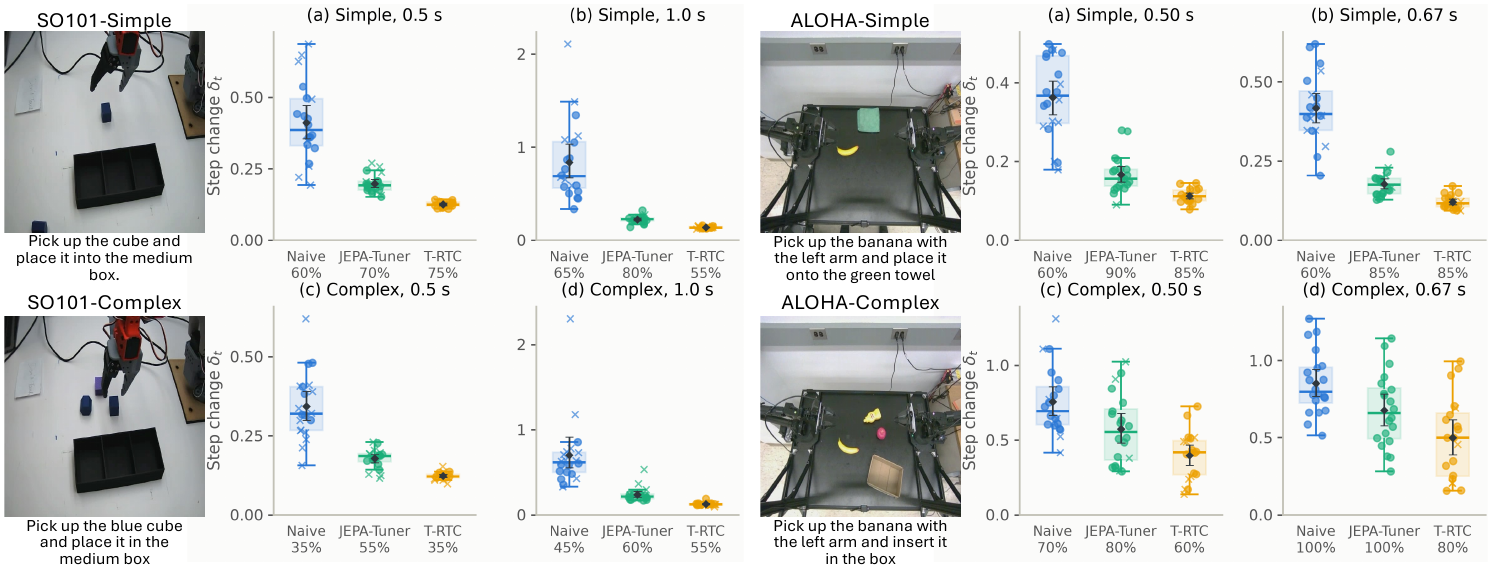}
    \vspace{-22pt}
    \caption{\textbf{Task success and action continuity in real-robot experiments.} SO-101 (left) and ALOHA (right) each show the simple task above the complex task. Each circle (success) or cross (failure) gives one trial's mean boundary step change. Percentages report success over 20 trials under the methods' names. Boxes show medians and interquartile ranges.}
    \label{fig:real_robot_async}
    \vspace{-15pt}
\end{figure*}
\vspace{-4pt}
\subsection{Asynchronous Execution on Varying Latency}
\label{sec:async_inference}
\vspace{-2pt}
We next test whether the predictor learned during fine-tuning can support asynchronous execution by anticipating guidance at the next chunk handoff. Across the four LIBERO suites, SmolVLA and $\pi_{0.5}$ generate $50$-action chunks while the robot executes previously committed actions. We compare JEPA-Tuner with Naive Async., Temporal Ensembling (TE)~\citep{act}, BID~\citep{bid}, RTC~\citep{rtc}, Training-Time RTC (T-RTC)~\citep{training_time_rtc}, Legato-uniform~\citep{legato}, VLASH~\citep{vlash}, and REMAC~\citep{remac} (Table~\ref{tab:libero_async_latency}). We follow the baselines' official repositories and match adaptation budgets and optimizer steps; training details are provided in the supplementary material. We impose inference delays of $\Delta\in\{5,10,15\}$ control steps. All methods execute $E=10$ actions per chunk for $\Delta\in\{5,10\}$ and $E=15$ for $\Delta=15$. 

For Naive Async., T-RTC, and JEPA-Tuner, we assess trajectory smoothness using the \emph{action step change} between consecutive executed commands,
\begin{equation}
    \delta_t
    = \left\|\widetilde{\mathbf a}_t - \widetilde{\mathbf a}_{t-1}\right\|_2,
    \label{eq:libero_step_change}
\end{equation}
where $\widetilde{\mathbf a}_t$ contains the normalized translation and rotation components of the command at step $t$, excluding the binary gripper command. At a handoff, $\delta_t$ measures the change from the last executed command of the preceding chunk to the first executed command of the new chunk. Smaller values indicate a smoother transition.

\begin{figure}[t]
    \centering
    \includegraphics[width=\columnwidth]{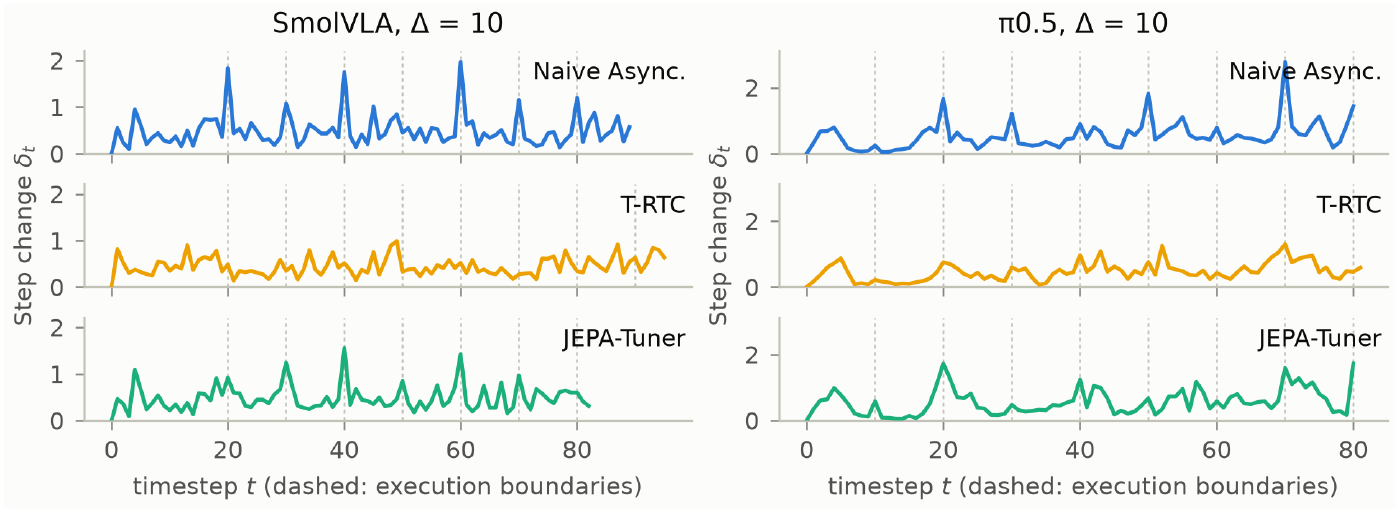}
    \vspace{-16pt}
    \caption{\textbf{Action-step changes during asynchronous execution at $\Delta=10$.} Each trace shows a successful LIBERO episode for SmolVLA (left) or $\pi_{0.5}$ (right). Dashed lines mark chunk execution boundaries. The step change $\delta_t$ measures the distance between consecutive normalized continuous actions.}
    \label{fig:async_smoothness}
    \vspace{-19pt}
\end{figure}

JEPA-Tuner achieves the highest success rate in five of the six backbone--delay settings, as shown in Table~\ref{tab:libero_async_latency}. At $\Delta=5$, BID leads on SmolVLA ($79.4\%$ versus $77.3\%$), and JEPA-Tuner leads on $\pi_{0.5}$ ($95.8\%$). At $\Delta=10$, JEPA-Tuner reaches $76.8\%$ on SmolVLA and $85.4\%$ on $\pi_{0.5}$, exceeding BID and REMAC by $4.5$ and $5.3\%$, respectively. At $\Delta=15$, where all methods also increase the execution horizon to $E=15$, JEPA-Tuner reaches $72.1\%$ and $58.2\%$, widening its lead to $10.8\%$ over VLASH on SmolVLA and $11.7\%$ over REMAC on $\pi_{0.5}$. At this delay, T-RTC and Legato reach $34.6\%$ and $34.1\%$ on $\pi_{0.5}$. Training-based approaches to asynchronous control, such as T-RTC and Legato, modify the diffusion or flow-matching probability path so that the action expert learns to generate a continuation consistent with the committed action prefix. At high latency, this prefix spans a larger fraction of the chunk and may dominate generation of the remaining actions, biasing the expert toward continuing the previous motion even when visual or proprioceptive cues call for an adjustment.

The successful rollouts in Fig.~\ref{fig:async_smoothness} illustrate the trade-off at $\Delta=10$. Naive Async.\ decodes from the observation available at query time and skips actions whose execution times have passed. Its traces show sharp changes at several chunk handoffs. T-RTC conditions each new chunk on the committed prefix and produces the smoothest transitions in these examples. JEPA-Tuner uses the committed actions to predict guidance at the handoff, allowing the robot to execute the first $E$ actions of the new chunk without overlap correction. Its traces show smaller spikes than Naive Async.\ at several handoffs, while its success rate exceeds T-RTC's by $15.7\%$ on SmolVLA and $11.0\%$ on $\pi_{0.5}$.

\begin{table}[t]
    \centering
    \caption{\textbf{Asynchronous success on LIBERO} (\%), averaged across the four suites. Column headers give $\Delta$ in control steps. Bold marks the best result in each column.}
    \label{tab:libero_async_latency}
    \footnotesize
    \setlength{\tabcolsep}{2pt}
    \renewcommand{\arraystretch}{1.05}
    \vspace{-11pt}
    \begin{tabular*}{\columnwidth}{@{\extracolsep{\fill}}lcccccc@{}}
        \toprule
        & \multicolumn{3}{c}{SmolVLA} & \multicolumn{3}{c}{$\pi_{0.5}$} \\
        \cmidrule(lr){2-4}\cmidrule(lr){5-7}
        Method & 5 & 10 & 15 & 5 & 10 & 15 \\
        \midrule
        Naive Async. & 77.6 & 68.5 & 53.8 & 94.5 & 73.5 & 32.4 \\
        TE~\citep{act} & 78.1 & 72.0 & 58.4 & 64.6 & 53.5 & 33.0 \\
        BID~\citep{bid} & \textbf{79.4} & 72.3 & 60.1 & 94.7 & 74.9 & 34.9 \\
        RTC~\citep{rtc} & 59.6 & 58.4 & 57.3 & 89.2 & 72.5 & 31.1 \\
        \midrule
        T-RTC~\citep{training_time_rtc} & 64.5 & 61.1 & 58.3 & 94.5 & 74.4 & 34.6 \\
        Legato~\citep{legato} & 45.8 & 43.1 & 42.5 & 89.2 & 67.0 & 34.1 \\
        VLASH~\citep{vlash} & 76.3 & 72.0 & 61.3 & 52.5 & 45.7 & 22.2 \\
        REMAC~\citep{remac} & 45.9 & 52.1 & 57.4 & 94.7 & 80.1 & 46.5 \\
        \textbf{JEPA-Tuner} & 77.3 & \textbf{76.8} & \textbf{72.1}
          & \textbf{95.8} & \textbf{85.4} & \textbf{58.2} \\
        \bottomrule
    \end{tabular*}
    \vspace{-12pt}
\end{table}

\begin{table}[t]
    \centering
    \caption{\textbf{Computational efficiency on LIBERO.} We report parameters used at inference (M) and generation time per chunk (ms, mean $\pm$ standard deviation).}
    \vspace{-10pt}
    \label{tab:inference_cost}
    \footnotesize
    \setlength{\tabcolsep}{1.5pt}
    \begin{tabular*}{\columnwidth}{@{\extracolsep{\fill}}lrrrr@{}}
        \toprule
        & \multicolumn{2}{c}{SmolVLA}
        & \multicolumn{2}{c}{$\pi_{0.5}$} \\
        \cmidrule(lr){2-3}\cmidrule(lr){4-5}
        Inference mode & Params (M) & Time (ms) & Params (M) & Time (ms) \\
        \midrule
        Standard FT (sync) & $450.0$ & $216.3 \pm 1.4$
                           & $4{,}143.4$ & $277.7 \pm 5.4$ \\
        JEPA-Tuner (sync)  & $454.2$ & $216.9 \pm 0.9$
                           & $4{,}161.2$ & $279.2 \pm 1.5$ \\
        JEPA-Tuner (async) & $478.1$ & $223.2 \pm 1.0$
                           & $4{,}250.1$ & $287.6 \pm 0.8$ \\
        \bottomrule
    \end{tabular*}
    \vspace{-20pt}
\end{table}
\vspace{-14pt}
\paragraph{Computational efficiency.}
To assess the computational cost of JEPA-Tuner's gains in task success and trajectory smoothness, we quantify the trainable parameters added during fine-tuning and measure inference time per action chunk. We compare three configurations on LIBERO, as shown in Table~\ref{tab:inference_cost}. Standard FT runs the base VLA alone. Synchronous JEPA-Tuner adds only the guidance adapter and disables the predictor because the action expert decodes from observed guidance. The adapter increases the parameter count by $0.9\%$ on SmolVLA and $0.4\%$ on $\pi_{0.5}$. The corresponding policies take only $0.3\%$ and $0.5\%$ longer per chunk than Standard FT, respectively, while achieving higher average success on LIBERO and Meta-World MT50. 

During asynchronous execution, JEPA-Tuner reuses the predictor learned during fine-tuning to anticipate guidance at the next chunk handoff. Together, the adapter and predictor increase the parameter count by $6.2\%$ on SmolVLA and $2.6\%$ on $\pi_{0.5}$, adding $6.9$ and $9.9$\,ms per chunk over Standard FT, respectively. However, we set the query lead to cover the end-to-end inference latency, including this overhead. So this increased inference time has a limited affect on real vla model deployment. Without further training or overlap correction, JEPA-Tuner reduced mean boundary step change relative to Naive Async.\ in all real-robot test, with competitive success rates, as shown in Table~\ref{tab:real_robot}.
\vspace{-2pt}
\subsection{Real-Robot Asynchronous Manipulation}
\label{sec:real_robot}

We evaluate the trade-off between task success and action trajectory smoothness on the small-size SO-101 and larger-size ALOHA using SmolVLA policies. Both SO-101 tasks require placing a cube in the medium compartment of a tray. In the complex task, the robot must select the blue cube beside a differently colored distractor. On ALOHA, the robot places a banana on a green towel or inserts it into a box. Across these tasks, we compare JEPA-Tuner with Naive Async.\ and T-RTC. To test how each method handles inference delay, we vary the query lead, the interval between requesting the next action chunk and its scheduled handoff. For real-world latency, we align JEPA-Tuner's prediction gap with a query lead that exceeds the maximum measured end-to-end response time. The details and video demos are shown in supplementary material.

We evaluate two deployment settings. In the local setting, SO-101 uses a $10$-step ($0.50$\,s) query lead, and ALOHA uses manually set leads of $15$ or $20$ steps ($0.50$ or $0.67$\,s). For remote inference, SO-101 in Minnesota queries a server in New York; the measured end-to-end latency is $0.8$--$0.9$\,s, so we use a $20$-step ($1.0$\,s) lead. Control frequencies and deployment details are given in the supplementary material. We run $20$ trials per method on each task and manually score whether the correct object reaches its destination. To measure action trajectory smoothness, we standardize joint and gripper targets using the task demonstrations and average the step change between the last executed target of one chunk and the first of the next across all action trajectories in each trial.

\begin{table}[t]
    \centering
    \caption{\textbf{Asynchronous inference on real-robots (SO-101 and ALOHA).}
    Success rate (\%, $\uparrow$) / mean boundary step change ($\downarrow$) over 20 trials per condition.
    Bold marks the best success rate per row, including ties.}
    \vspace{-10pt}
    \label{tab:real_robot}
    \footnotesize
    \setlength{\tabcolsep}{2pt}
    \renewcommand{\arraystretch}{0.95}
    \begin{tabular*}{\columnwidth}{@{\extracolsep{\fill}}lcccc@{}}
        \toprule
        Task (robot) & Query Lead (s) & Naive Async. & T-RTC & JEPA-Tuner \\
        \midrule
        Simple          & 0.50 & 60 / 0.411 & \textbf{75} / 0.125 & 70 / 0.198 \\
        \emph{(SO-101)} & 1.00 & 65 / 0.837 & 55 / 0.137 & \textbf{80} / 0.223 \\
        \addlinespace[1pt]
        Complex         & 0.50 & 35 / 0.343 & 35 / 0.123 & \textbf{55} / 0.178 \\
        \emph{(SO-101)} & 1.00 & 45 / 0.700 & 55 / 0.124 & \textbf{60} / 0.238 \\
        \midrule
        Simple          & 0.50 & 60 / 0.362 & 85 / 0.112 & \textbf{90} / 0.167 \\
        \emph{(ALOHA)}  & 0.67 & 60 / 0.417 & \textbf{85} / 0.120 & \textbf{85} / 0.177 \\
        \addlinespace[1pt]
        Complex         & 0.50 & 70 / 0.756 & 60 / 0.397 & \textbf{80} / 0.575 \\
        \emph{(ALOHA)}  & 0.67 & \textbf{100} / 0.851 & 80 / 0.498 & \textbf{100} / 0.677 \\
        \bottomrule
    \end{tabular*}
    \vspace{-20pt}
\end{table}

JEPA-Tuner achieves the highest observed success rate outright in five of the eight task--lead conditions and ties for highest in two (Table~\ref{tab:real_robot}). With high-latency on SO-101, JEPA-Tuner achieves $80\%$ success on the simple task and $60\%$ on the complex task, compared with $65\%$ and $45\%$, respectively, for Naive Async. On ALOHA simple tasks at $0.50$\,s, JEPA-Tuner achieves $90\%$ success, compared with $60\%$ for Naive Async.\ and $85\%$ for T-RTC.

Figure~\ref{fig:real_robot_async} relates task success to changes in commanded actions across all SO-101 and ALOHA chunk trajectories. JEPA-Tuner reduces the mean boundary step change relative to Naive Async.\ in every condition, by $48.1$--$73.4\%$ on SO-101 and $20.4$--$57.6\%$ on ALOHA. T-RTC yields the smallest boundary changes in all eight conditions. JEPA-Tuner achieves higher success than T-RTC in six conditions, ties in one, and trails only on the simple SO-101 task at $0.50$\,s ($70\%$ versus $75\%$). On ALOHA insertion at $0.67$\,s, JEPA-Tuner matches the $100\%$ success rate of Naive Async.\ while reducing the mean boundary step change from $0.851$ to $0.677$. T-RTC reduces the mean change further to $0.498$ and achieves $80\%$ success. Conditioning each new chunk on the committed action prefix, as in T-RTC and Legato, encourages continuity across handoffs. The LIBERO traces and real-robot results illustrate this benefit for T-RTC, but its lower task success suggests that a strong bias toward continuing the previous motion may limit the policy's ability to adjust when the task calls for a change. JEPA-Tuner uses the committed actions to predict handoff guidance, allowing the action expert to generate a new chunk for the anticipated state without overlap correction. It achieves higher success than both T-RTC and Legato in all six LIBERO backbone--delay settings (Table~\ref{tab:libero_async_latency}). On the real robots, it reduces mean boundary step change relative to Naive Async.\ in all eight conditions while matching or exceeding the best baseline success rate in seven (Table~\ref{tab:real_robot}).

%% file: sec/6_Conclusions.tex
\vspace{-6pt}
\section{Conclusions}
\label{sec:Conclusions}
\vspace{-2pt}
In this paper, we identify two challenges in VLA learning and deployment: underexploited action-conditioned dependencies in multimodal guidance and stale conditioning during asynchronous execution. Motivated by them, we introduce a unified VLA fine-tuning framework, \textbf{JEPA-Tuner}. During fine-tuning, joint supervision of future-guidance prediction and action generation captures temporal dependencies in the VLA's native guidance while preserving its utility for control. During asynchronous execution, the same predictor uses committed actions to anticipate guidance at the next handoff, enabling each new chunk to execute from its first action without overlap correction. Experiments on LIBERO and Meta-World MT50 demonstrate improved fine-tuning performance across all three evaluated backbones, even with the predictor inactive. Asynchronous evaluations in simulation and on SO-101 and ALOHA further support a favorable trade-off between task success and action continuity under inference latency. Although JEPA-Tuner has the potential to extend to large-scale VLA pretraining and more diverse real-world tasks, our evaluation remains limited to small-scale fine-tuning and two robot platforms; we leave these extensions to future work.

%% file: sec/7_LLM_Usage_Statement.tex
\section{LLM Usage Statement}
\label{sec:Conclusions}

In this work, we used LLMs (ChatGPT and Claude Code) to assist with paper writing and method implementation. We did not use these tools to generate research ideas, develop theoretical models, design research methodology or experiments. We have reviewed all content associated with this work, including LLM-assisted text and code, for accuracy, correctness, and consistency with our intended methodology. We take full responsibility for the final content of this work, including all text, claims, and artifacts produced with the assistance of LLMs.

%% file: sec/JEPA_Tuner_Supplement.tex

\section*{Table of Contents}
\begingroup
\small
\setlength{\parskip}{0pt}
\newcommand{\suppcontentsentry}[3]{%
    \noindent\hspace*{#1}\makebox[2.5em][l]{\ref{#2}}%
    #3\nobreak\dotfill\pageref{#2}\par}
\suppcontentsentry{0em}{sec:supp_ablations}{\textbf{Ablation Studies}}
\suppcontentsentry{1em}{sec:supp_ablation_prediction}{Future-guidance prediction}
\suppcontentsentry{1em}{sec:supp_ablation_policy}{Policy without prediction}
\suppcontentsentry{0em}{sec:supp_additional_results}{\textbf{Additional Experimental Results}}
\suppcontentsentry{0em}{sec:supp_jepa_implementation}{\textbf{Implementation Details of JEPA-Tuner}}
\suppcontentsentry{0em}{sec:supp_experimental_settings}{\textbf{Experimental Settings}}
\suppcontentsentry{1em}{sec:supp_sync_simulation}{Synchronous simulation benchmarks}
\suppcontentsentry{1em}{sec:supp_async_simulation}{Asynchronous simulation and latency}
\suppcontentsentry{1em}{sec:supp_physical_protocols}{Physical manipulation protocols}
\suppcontentsentry{1em}{sec:supp_action_continuity}{Action continuity and aggregation}
\endgroup

\section{Ablation Studies}
\label{sec:supp_ablations}

We conduct two ablation studies on Meta-World MT50~\citep{metaworld} with SmolVLA and $\pi_{0.5}$. To test whether the action-conditioned predictor anticipates the guidance needed at the next chunk handoff, we evaluate the predicted end-effector position and the actions decoded from the predicted guidance at the scheduled execution steps. To test whether self-predictive fine-tuning improves task success from observed multimodal guidance, we disable the predictor at inference and compare JEPA-Tuner with Standard FT under synchronous inference and a shared naive asynchronous schedule.

\subsection{Future-guidance prediction}
\label{sec:supp_ablation_prediction}

We evaluate ten-step prediction on held-out frames from 50 demonstration episodes. For an observation at step $t$, we compare the ten actions that both methods would execute from $t+10$ through $t+19$. JEPA-Tuner decodes from predicted guidance for $t+10$ and uses chunk entries $0$--$9$. Naive asynchronous inference decodes from the observation at $t$ and skips the first ten entries, using entries $10$--$19$. We use the same action expert for all three decodes, taking the first ten actions decoded from the observed guidance at $t+10$ as the reference. For end-effector position error, we compare JEPA-Tuner's predicted position and the position at $t$ used by the naive baseline with the observed position at $t+10$.

Table~\ref{tab:supp_ablation_prediction} shows that JEPA-Tuner reduces relative executed-action error by $26.4\%$ on SmolVLA and $27.2\%$ on $\pi_{0.5}$ compared with naive asynchronous inference. It also reduces the end-effector position error from $63.2$\,mm to $6.7$\,mm and $5.0$\,mm, respectively. These results show that predicting guidance helps the action expert generate actions closer to those conditioned on the handoff observation.

\begin{table}[!ht]
    \centering
    \caption{\textbf{Time-aligned prediction accuracy on Meta-World MT50.} Pred. uses guidance and state predicted for $t+10$. Naive decodes from the observation at $t$ and skips the first ten actions.}
    \label{tab:supp_ablation_prediction}
    \footnotesize
    \setlength{\tabcolsep}{2pt}
    \begin{tabular*}{\columnwidth}{@{\extracolsep{\fill}}lrrrr@{}}
        \toprule
        & \multicolumn{2}{c}{SmolVLA} & \multicolumn{2}{c}{$\pi_{0.5}$} \\
        \cmidrule(lr){2-3} \cmidrule(lr){4-5}
        Metric & Pred. & Naive & Pred. & Naive \\
        \midrule
        Relative executed-action error $\downarrow$ & \textbf{0.192} & 0.261 & \textbf{0.155} & 0.213 \\
        End-effector position error (mm) $\downarrow$ & \textbf{6.7} & 63.2 & \textbf{5.0} & 63.2 \\
        \bottomrule
    \end{tabular*}
\end{table}

\subsection{Policy without prediction}
\label{sec:supp_ablation_policy}

We disable the predictor at inference for the jointly trained JEPA-Tuner checkpoint and compare its task success with Standard FT. We evaluate both policies synchronously and under the same naive asynchronous schedule with a delay of $\Delta=10$ control steps. Each condition comprises 50 episodes on each of the 50 tasks, for 2,500 episodes. Table~\ref{tab:supp_ablation_predictor_off} shows that JEPA-Tuner improves synchronous success by $1.6$ percentage points on SmolVLA and $6.8$ points on $\pi_{0.5}$. Under the matched asynchronous schedule, the gains are $1.6$ and $3.9$ points, respectively.

\begin{table}[!ht]
    \centering
    \caption{\textbf{Success rates on Meta-World MT50 (\%).} JEPA denotes JEPA-Tuner with the predictor disabled in both settings.}
    \label{tab:supp_ablation_predictor_off}
    \footnotesize
    \setlength{\tabcolsep}{2pt}
    \begin{tabular*}{\columnwidth}{@{\extracolsep{\fill}}lrrrr@{}}
        \toprule
        & \multicolumn{2}{c}{SmolVLA} & \multicolumn{2}{c}{$\pi_{0.5}$} \\
        \cmidrule(lr){2-3} \cmidrule(lr){4-5}
        Inference & Std. FT & JEPA & Std. FT & JEPA \\
        \midrule
        Synchronous & 64.4 & \textbf{66.0} & 37.2 & \textbf{44.0} \\
        Naive async. ($\Delta=10$) & 58.1 & \textbf{59.7} & 35.0 & \textbf{38.9} \\
        \bottomrule
    \end{tabular*}
\end{table}

\section{Additional Experimental Results}
\label{sec:supp_additional_results}

To complement the individual LIBERO rollouts in Fig.~4 of the main paper, we compare the action step change $\delta_t$ at chunk handoffs (Eq.~(12) of the main paper) across Naive Async., T-RTC, and JEPA-Tuner. We use chunk size $H=50$ and a nominal execution horizon of $E=10$ at delays $\Delta\in\{5,10\}$ control steps. On SmolVLA at $\Delta=10$, the initial chunk executes $E+\Delta$ actions before the first handoff. For each method, backbone, and delay, we trace 400 episodes with matched initializations across methods, using ten episodes per task in each of the four LIBERO suites. We average $\delta_t$ over all valid handoffs across these episodes, as described in Sec.~\ref{sec:supp_action_continuity}. Action normalization differs across backbones, so we compare methods within each backbone.

\begin{table}[!ht]
    \centering
    \caption{\textbf{Mean action step change at chunk handoffs on LIBERO.} Each entry reports mean $\delta_t$ ($\downarrow$) during asynchronous execution. Bold marks the lowest value in each setting.}
    \label{tab:supp_libero_step_change}
    \footnotesize
    \setlength{\tabcolsep}{2pt}
    \begin{tabular*}{\columnwidth}{@{\extracolsep{\fill}}lcccc@{}}
        \toprule
        & \multicolumn{2}{c}{SmolVLA} & \multicolumn{2}{c}{$\pi_{0.5}$} \\
        \cmidrule(lr){2-3} \cmidrule(lr){4-5}
        Method & $\Delta=5$ & $\Delta=10$ & $\Delta=5$ & $\Delta=10$ \\
        \midrule
        Naive Async. & 1.13 & 1.23 & 0.89 & 1.37 \\
        T-RTC & \textbf{0.38} & \textbf{0.38} & \textbf{0.43} & \textbf{0.45} \\
        JEPA-Tuner & 1.26 & 1.30 & 0.77 & 1.13 \\
        \bottomrule
    \end{tabular*}
\end{table}

Table~\ref{tab:supp_libero_step_change} shows that on $\pi_{0.5}$, JEPA-Tuner reduces the mean handoff step change relative to Naive Async.\ from $0.89$ to $0.77$ at $\Delta=5$ and from $1.37$ to $1.13$ at $\Delta=10$. On SmolVLA, the pooled means are slightly higher for JEPA-Tuner at both delays. The SmolVLA rollout in Fig.~4 of the main paper illustrates smaller step changes at several individual handoffs. T-RTC produces the smallest mean handoff step change in all four settings, consistent with its use of committed-prefix conditioning. JEPA-Tuner uses the committed actions to predict guidance at the handoff and executes the first $E$ actions of the new chunk without overlap correction. In the traced rollouts, JEPA-Tuner achieves higher task success than T-RTC in every setting and exceeds Naive Async.\ on both backbones at $\Delta=10$, consistent with Table~3 of the main paper. These results illustrate the trade-off between handoff continuity and task success.

\section{Implementation Details of JEPA-Tuner}
\label{sec:supp_jepa_implementation}

We freeze the pretrained vision--language backbone and jointly train a cross-modal guidance adapter, an action-conditioned predictor, the action expert, and any trainable state projection. For a valid pair of observations $d$ steps apart in the same demonstration, the predictor takes the adapted guidance $g_t$, measured state $s_t$, intervening demonstrated actions $a_t,\ldots,a_{t+d-1}$, and gap $d$. Its targets are the visual guidance encoded from the later observation and the measured state $s_{t+d}$. The task instruction stays fixed across the pair. We train the same action expert to generate the demonstrated chunk starting at $t$ from observed guidance and the chunk starting at $t+d$ from predicted guidance. Both terms use the backbone's flow-matching objective, so the predictor must encode information needed for the actions that follow the predicted state. Synchronous execution uses observed guidance. Asynchronous execution reuses the predictor with the actions already committed through the next handoff.

For SmolVLA and $\pi_{0.5}$~\citep{smolvla,pi05}, we insert a one-layer residual attention adapter at the input prefix. It has eight attention heads and a feed-forward bottleneck of 256. Zero-initialized output projections make the adapter an identity map at initialization. The predictor is a two-layer transformer with eight heads and feed-forward width 2,048. It reads the adapted prefix, eight tokens encoding the intervening actions, and one learned gap token. Residual heads predict changes to the visual embeddings and the normalized state. The loss weights for the current action loss, normalized visual squared error, visual cosine distance, normalized state squared error, and future action loss are $1$, $1$, $0.1$, $1$, and $1$, respectively. Visual and state errors are normalized separately. We omit padding and fixed instruction tokens from the prediction targets.

The two backbones encode proprioception differently. SmolVLA projects the robot state into one token. On LIBERO, its prefix has 177 positions of width 960, comprising 64 tokens per camera for two cameras, 48 instruction positions, and the state token. We predict proprioception in normalized state coordinates, pass the prediction through SmolVLA's trainable state projection, and insert it at the state position. We run the adapter on this prefix to recover the state token in context, then combine it with the predicted visual tokens and the observed instruction. This preserves the state input used by the action expert.

In $\pi_{0.5}$, the state enters through the instruction prompt. On LIBERO, the PaliGemma prefix has 968 positions of width 2,048. Each of three camera slots occupies 256 positions, with two real views and one masked slot, followed by 200 prompt positions. We discretize the predicted state into 256 bins, rebuild the prompt with the same instruction, update its padding mask, and apply the adapter in context. We combine the rebuilt prompt with the predicted image embeddings. Each valid camera contributes equally to the normalized visual prediction loss.

For the Meta-World experiments labeled ``GR00T'' in Table~2 of the main paper, we use the GR00T-N1.7-3B checkpoint~\citep{gr00t}, with its backbone frozen and its action head configured for the new embodiment. A one-layer, 16-head cross-modal adapter includes a state token in the features used for action conditioning. The action-conditioned predictor has approximately 110 million parameters. We train the action head on predicted guidance with a future action loss weight of $1.0$, and use the observed guidance for the synchronous evaluation. Both Standard FT and JEPA-Tuner use 20-action chunks and execute ten actions per chunk on this benchmark.

\section{Experimental Settings}
\label{sec:supp_experimental_settings}

For every comparison, we match the training budget within each backbone: all methods that we train build on the same pretrained weights and demonstrations, with no baseline receiving fewer optimizer steps than JEPA-Tuner, and the training-free asynchronous baselines reuse the Standard FT checkpoint.

\subsection{Synchronous simulation benchmarks}
\label{sec:supp_sync_simulation}
Standard FT trains the action expert to generate demonstrated actions from the current observation. On LIBERO~\citep{libero}, we evaluate the Spatial, Object, Goal, and LIBERO-10 suites. A rollout succeeds when the environment's task-specific goal condition is satisfied within its episode limit. The suite success rate is the fraction of successful rollouts, and the LIBERO average gives each suite equal weight. The $\pi_{0.5}$ and GR00T Standard FT rows use published backbone results~\citep{vla_jepa}, while we obtain the SmolVLA Standard FT row by fine-tuning its pretrained base following the official configuration. For synchronous inference, JEPA-Tuner uses the current observation directly with no imposed delay. The JEPA-Tuner checkpoints for SmolVLA and $\pi_{0.5}$ used in these LIBERO evaluations were trained with gap $d=5$.

On Meta-World MT50~\citep{metaworld}, we use 50 scripted demonstrations per task. The dataset has one $480\times480$ camera view and four-dimensional state and action vectors. We evaluate 50 episodes per task with the same episode across policies and a 500-step cap. An episode counts as a success if the environment raises its task-success flag within this limit. Easy, Medium, Hard, and Very Hard contain 28, 11, 6, and 5 tasks, contributing 1,400, 550, 300, and 250 episodes, respectively. We divide the total number of successful episodes by 2,500 to obtain the overall rate, which weights each difficulty group by its episode count.

We match the demonstrations and training budget within each backbone. SmolVLA and GR00T train for 20,000 optimizer steps with effective batch size 64. SmolVLA uses float32. GR00T uses bfloat16. The $\pi_{0.5}$ configuration uses 30,000 steps, effective batch size 28, and bfloat16. For these Meta-World experiments, JEPA-Tuner uses a training gap of $d=10$ for all three backbones. SmolVLA and $\pi_{0.5}$ decode 50-action chunks and GR00T decodes 20-action chunks, with ten actions executed per chunk in each case.

\subsection{Asynchronous simulation and latency accounting}
\label{sec:supp_async_simulation}
Let $H$ be the number of decoded actions, $E$ the number executed per chunk, and $b$ the control step at which the next chunk is scheduled to start. A single worker issues the query at step $q=b-\Delta$ while the previous chunk supplies the committed actions through step $b-1$. For nominal control frequency $f$ in Hz, the scheduled observation-to-handoff interval is
\begin{equation}
    \tau_{\mathrm{lead}}=\frac{b-q}{f}=\frac{\Delta}{f}.
    \label{eq:supp_query_lead}
\end{equation}
The response time for an individual request is defined on the robot client's clock as
\begin{equation}
    \tau_{\mathrm{resp}}=T_{\mathrm{ready}}-T_{\mathrm{start}},
    \label{eq:supp_response_time}
\end{equation}
where $T_{\mathrm{start}}$ marks the start of the client request and $T_{\mathrm{ready}}$ marks the availability of the returned action chunk. This interval includes transmission, remote processing, and return of the actions. Sensor age at request time and any wait between receipt and the scheduled handoff are separate intervals. An early response can wait until $b$, so the scheduled lead and observed response time can differ.

For variable real-world latency, we profile the complete request path, including JEPA-Tuner's prediction overhead, and choose a query lead above the maximum measured response time, rather than the mean. If $\tau_{\max}$ is this measured maximum and $\varepsilon>0$ is a margin in seconds, we choose
\begin{equation}
    d=\Delta=\left\lceil f(\tau_{\max}+\varepsilon)\right\rceil,
    \qquad \frac{\Delta}{f}>\tau_{\max}.
    \label{eq:supp_latency_margin}
\end{equation}
The prediction gap $d$ therefore targets the scheduled handoff, rather than the response time of an individual request. For example, if a cloud deployment has typical response times near $0.6$\,s and a measured maximum of $0.9$\,s, we can use a $1.0$\,s lead, corresponding to $d=\Delta=20$ at $20$\,Hz. An early response is buffered until the scheduled handoff, and the previous chunk supplies the committed actions until then. We use $E\geq\Delta$ for a single sequential worker. Response times beyond the profiled bound are handled by the late-arrival policy described below.

In LIBERO, we impose the delay in control steps by making a request issued at $b-\Delta$ available at $b$, independently of host inference speed. We use $H=50$ throughout. For $\Delta=5,10,15$, the execution horizons are $E=10,10,15$, respectively. At the high-latency setting, the longer horizon allows one worker to complete each request before the next one starts. Comparisons across those settings therefore change both delay and execution horizon. JEPA-Tuner uses a separately fine-tuned checkpoint with $d=\Delta$ for each delay.

Following FutureRTC~\citep{futurertc}, Naive Async., Temporal Ensembling (TE)~\citep{act}, BID~\citep{bid}, and RTC~\citep{rtc} use the same conventionally fine-tuned checkpoint for a given backbone. Training-Time RTC (T-RTC)~\citep{training_time_rtc}, Legato~\citep{legato}, VLASH~\citep{vlash}, and REMAC~\citep{remac} use the same LIBERO demonstrations with their respective training objectives. We adapt the official implementations to the common VLA interfaces and match the additional adaptation budgets and optimizer steps within each backbone. These step counts exclude any prior Standard FT stage. Table~\ref{tab:supp_async_initialization} gives the checkpoint initializations and adaptation budgets for T-RTC and Legato.

\begin{table}[t]
    \centering
    \caption{\textbf{Checkpoint initialization and adaptation on LIBERO.} Standard FT initializations have already received 20,000 steps on SmolVLA and 30,000 on $\pi_{0.5}$. Legato uses effective batch size 32.}
    \label{tab:supp_async_initialization}
    \footnotesize
    \setlength{\tabcolsep}{2pt}
    \begin{tabular*}{\columnwidth}{@{\extracolsep{\fill}}lllr@{}}
        \toprule
        Backbone & Method & Initial weights & Extra steps \\
        \midrule
        SmolVLA & T-RTC & Pretrained base & 20,000 \\
                & Legato & Pretrained base & 20,000 \\
        $\pi_{0.5}$ & T-RTC & Standard FT & 10,000 \\
                   & Legato & Standard FT & 10,000 \\
        \bottomrule
    \end{tabular*}
\end{table}

For the Legato results reported as ``Legato-uniform'' in Sec.~5.2 of the main paper (abbreviated to ``Legato'' in Table~3), we use the paper's uniformly sampled training delays, extending the range to cover $0$--$15$ control steps. The guided-prefix length $d_{\mathrm L}$ is sampled as
\begin{equation}
    \Pr(d_{\mathrm L}=k)=\frac{1}{16},
    \qquad k\in\{0,\ldots,15\}.
    \label{eq:supp_legato_uniform}
\end{equation}
The training ramp length is zero with probability $0.5$ and otherwise sampled uniformly from $1$ to $35$ control steps. SmolVLA starts from its pretrained base and $\pi_{0.5}$ from its standard LIBERO fine-tune, as detailed in Table~\ref{tab:supp_async_initialization}. We train the action expert with the vision--language backbone frozen. At deployment, SmolVLA uses a linear ramp of length $r=H-\Delta-E$, while $\pi_{0.5}$ uses a step schedule with no ramp. In both cases, the first $\Delta$ actions are fully guided by the committed prefix. Legato uses five Euler denoising steps, whereas the other asynchronous rows use ten. Each reported latency setting is evaluated over 2,000 episodes across the four suites.

\paragraph{Choice of delay sampling.}
The official simulation code also uses an exponential guided-prefix sampler, which, when rescaled to our delay range, gives
\begin{equation}
    \Pr(d_{\mathrm L}=k)
    =\frac{\exp(-4k/15)}{\sum_{j=0}^{15}\exp(-4j/15)},
    \qquad k\in\{0,\ldots,15\}.
    \label{eq:supp_legato_exponential}
\end{equation}
Under this distribution, prefix lengths $10$ and $15$ occur in approximately $1.65\%$ and $0.43\%$ of training samples, respectively, compared with $6.25\%$ each under uniform sampling. Longer delays therefore receive substantially less training exposure. We retain uniform sampling to ensure equal sampling probability across the evaluated delay range, including the larger gaps relevant to physical deployment with substantial or variable inference latency.

Naive Async.\ skips the actions whose intended times have elapsed. T-RTC and Legato condition the new chunk on the committed prefix and execute the remaining suffix in the execution horizon. JEPA-Tuner decodes from predicted boundary guidance, so an on-time handoff starts at the first action of the new chunk.

\subsection{Physical manipulation protocols}
\label{sec:supp_physical_protocols}
Table~\ref{tab:supp_robot_timing} gives the four physical deployment settings. On SO-101, the robot and control client are in Minnesota. The shorter-lead condition uses a Minnesota server, and the longer-lead condition sends observations to an NVIDIA RTX A6000 server in New York. The approximate end-to-end response time for the New York deployment is $0.8$--$0.9$\,s, equivalent to roughly 16--18 control steps at $20$\,Hz. We schedule handoff after 20 steps, giving a $1.0$\,s nominal lead. On ALOHA, transmitting four camera views to a distant server makes tight deadlines difficult to meet. We run inference locally and manually set the query lead with no additional artificial sleep. At the nominal $30$\,Hz control rate, 15 steps give exactly $0.50$\,s and 20 steps give $2/3$\,s, displayed as $0.67$\,s.

\begin{table}[t]
    \centering
    \caption{\textbf{Physical deployment timing.} All methods decode $H=50$ actions. The lead is the scheduled time from query to handoff, computed as $\Delta/f$.}
    \label{tab:supp_robot_timing}
    \footnotesize
    \setlength{\tabcolsep}{2pt}
    \begin{tabular*}{\columnwidth}{@{\extracolsep{\fill}}llcccc@{}}
        \toprule
        Robot & Inference site & $f$ (Hz) & $\Delta$ & $E$ & Lead (s) \\
        \midrule
        SO-101 & Minnesota & 20 & 10 & 10 & 0.50 \\
               & New York  & 20 & 20 & 20 & 1.00 \\
        ALOHA  & Local     & 30 & 15 & 25 & 0.50 \\
               & Local     & 30 & 20 & 25 & 0.67 \\
        \bottomrule
    \end{tabular*}
\end{table}

On SO-101, the simple task requires placing a cube in the medium compartment of a tray. The complex task adds a differently colored distractor and requires placing the blue cube in that compartment. Each task has 50 demonstrations. The policy receives two $640\times480$ overhead views, one $320\times240$ wrist view, the instruction, and six state coordinates. The client transmits JPEG images and executes absolute targets for five arm joints and the gripper. The policy runs at $20$\,Hz, and the local motor loop runs at a nominal $60$\,Hz. The Minnesota and New York deployments use different servers, transports, and execution horizons.

The robot holds its starting pose for the first chunk. During execution, an on-time JEPA-Tuner response starts at index zero, whereas Naive Async.\ and T-RTC start at index $\Delta$. If $k$ additional policy steps elapse before a late chunk can be installed, the corresponding start indices advance to $k$ and $\Delta+k$. In sessions using the deadline-aware controller, the robot holds its position when a chunk exceeds its $0.25$\,s late-arrival tolerance. We retain all 20 episodes for each method and task.

For the SO-101 runs at $d=20$, we use 48 demonstrations from each task for training and hold out two for offline evaluation. All three methods use the same SmolVLA base, 10,000 optimizer steps, batch size 32, and float32 on one RTX A6000. AdamW uses a peak learning rate of $10^{-4}$, 1,000 warmup steps, cosine decay to $2.5\times10^{-6}$, and gradient clipping at 10. The JEPA-Tuner runs use state-space prediction, camera-wise visual normalization, future action supervision, and an auxiliary-to-action loss cap of $0.5$. For each physical lead, the JEPA-Tuner prediction gap and T-RTC maximum training delay match $\Delta$.

On ALOHA, both tasks use the left arm. The simple task requires picking up a banana and placing it on a green towel. The complex task requires inserting the banana into a box. The input comprises high and low external views and both wrist views, all $640\times480$ RGB. The logged action interface contains six joint targets and one gripper target per arm. In the simple task, the seven predicted left-arm targets are expanded to this interface with the right arm held. The demonstrations were recorded at $50$\,Hz, and deployment uses a nominal $30$\,Hz with $E=25$. At an on-time handoff, JEPA-Tuner starts at action zero and Naive Async.\ and T-RTC start at action $\Delta$.

On both platforms, we manually label success when the correct object reaches the instructed destination. The two tasks, two leads, and three methods give 240 scored trials per platform. 

\subsection{Action continuity and aggregation}
\label{sec:supp_action_continuity}
On LIBERO, we use the action step change $\delta_t$ defined in Eq.~(12) of the main paper. We compute it over the six normalized translation and rotation coordinates of consecutive executed actions, excluding the binary gripper command. A handoff pairs the last executed action of one chunk with the first executed action of the next. For each backbone and delay, we pool handoff and within-chunk changes separately over 400 episodes with matched initializations across methods. Only valid consecutive pairs before episode termination contribute. The LIBERO traces in the main paper show individual successful episodes.

For SO-101 and ALOHA, we retain the Euclidean step-change definition $\delta_t$ but apply it to the absolute policy targets selected before local speed limiting. Unlike the six continuous LIBERO coordinates, the physical-robot targets include the joint and gripper coordinates. We standardize each coordinate using its task-specific demonstration mean and standard deviation, shared across methods and query leads. The target includes all six joint and gripper coordinates on SO-101 and all 14 on ALOHA. At the first control step $b$ of a newly installed chunk, the handoff contributes
\begin{equation}
    \delta_b =
        \left\|\widetilde{\mathbf a}_{b}^{\,\mathrm{new}}
             -\widetilde{\mathbf a}_{b-1}^{\,\mathrm{old}}\right\|_2,
    \label{eq:action_step_change}
\end{equation}
where the superscripts identify the chunks that supplied the two successive targets.